\documentclass{article}

\usepackage{microtype}
\usepackage{graphicx}
\usepackage{subfigure}
\usepackage{booktabs} 
\usepackage{xcolor}
\usepackage{booktabs}
\usepackage{colortbl}
\usepackage{array}
\usepackage{multirow}

\usepackage{hyperref}
\usepackage{tcolorbox}
\usepackage{tikz}
\usetikzlibrary{calc}
\usetikzlibrary{arrows.meta, positioning, bending}

\usepackage[accepted]{icml2025}

\usepackage{amsmath}
\usepackage{amssymb}
\usepackage{mathtools}
\usepackage{amsthm}

\usepackage[capitalize,noabbrev]{cleveref}

\theoremstyle{plain}
\newtheorem{theorem}{Theorem}[section]

\theoremstyle{definition}
\newtheorem{definition}[theorem]{Definition}

\theoremstyle{remark}

\usepackage[textsize=tiny]{todonotes}

\icmltitlerunning{Language Games as the Pathway to Artificial Superhuman Intelligence}

\begin{document}

\twocolumn[
\icmltitle{Language Games as the Pathway to  Artificial Superhuman Intelligence}



\icmlsetsymbol{equal}{*}

\begin{icmlauthorlist}
\icmlauthor{Ying Wen}{sjtu}
\icmlauthor{Ziyu Wan}{sjtu}
\icmlauthor{Shao Zhang}{sjtu}
\end{icmlauthorlist}

\icmlaffiliation{sjtu}{Shanghai Jiao Tong University}

\icmlcorrespondingauthor{Ying Wen}{ying.wen@sjtu.edu.cn}

\icmlkeywords{Machine Learning, Language Games，Superhuman Intelligence}

\vskip 0.3in
]



\printAffiliationsAndNotice{}  

\begin{abstract}
The evolution of large language models (LLMs) toward artificial superhuman intelligence (ASI) hinges on data reproduction, a cyclical process in which models generate, curate and retrain on novel data to refine capabilities. Current methods, however, risk getting stuck in a data reproduction trap: optimizing outputs within fixed human-generated distributions in a closed loop leads to stagnation, as models merely recombine existing knowledge rather than explore new frontiers. In this paper, we propose language games as a pathway to expanded data reproduction, breaking this cycle through three mechanisms: (1) \textit{role fluidity}, which enhances data diversity and coverage by enabling multi-agent systems to dynamically shift roles across tasks; (2) \textit{reward variety}, embedding multiple feedback criteria that can drive complex intelligent behaviors; and (3) \textit{rule plasticity}, iteratively evolving interaction constraints to foster learnability, thereby injecting continual novelty. By scaling language games into global sociotechnical ecosystems, human-AI co-evolution generates unbounded data streams that drive open-ended exploration. This framework redefines data reproduction not as a closed loop but as an engine for superhuman intelligence.
\end{abstract}

\section{Introduction}
\label{sec:introduction}

The evolution of large language models (LLMs) on the pathway to artificial superhuman intelligence (ASI)~\citep{morris2023artificial} is fundamentally driven by \textbf{data reproduction}—an iterative process where models acquire novel data streams, evaluate outputs against human feedback and task-specific ground truth, and refine capabilities through targeted retraining~\citep{ouyang2022training}. This cyclical regeneration of knowledge creates a dynamic interaction between model improvement and the data ecosystem, where each iteration aims to better approximate real-world linguistic patterns~\citep{bommasani2021opportunities}. Unlike traditional machine learning paradigms constrained by fixed training sets, modern LLMs exhibit metabolic characteristics: continuously ingesting diverse inputs, transforming them through neural computations, and regenerating outputs that fuel subsequent learning cycles~\citep{thompson2007mind}.

The current data production paradigms widely used in large-scale pretraining~\citep{brown2020language}, supervised fine-tuning, and reinforcement learning from human feedback~\citep{christiano2017deep,luong2024reft} increasingly reveal their limitations. These methods, while effective for enhancing performance in established tasks, inadvertently confine the models to a \textit{data reproduction trap}. By optimizing outputs within fixed distributions of human-annotated examples and user preferences, they reinforce historical biases~\citep{shumailov2023curse} and prioritize short-term alignment over long-term intellectual growth. The result is a self-referential loop in which models become adept at recombining known concepts but struggle to generate fundamentally novel ideas: a stagnation akin to Marx's critique of economic systems trapped in simple reproduction \citep{marx_capital}. This impasse demands a paradigm shift from closed-loop optimization to open-ended conceptual exploration, ultimately pushing toward ASI by exploring beyond established distributional boundaries.

\begin{figure*}[ht!]
\vspace{-8pt}
    \centering
    \includegraphics[width=1.0\textwidth]{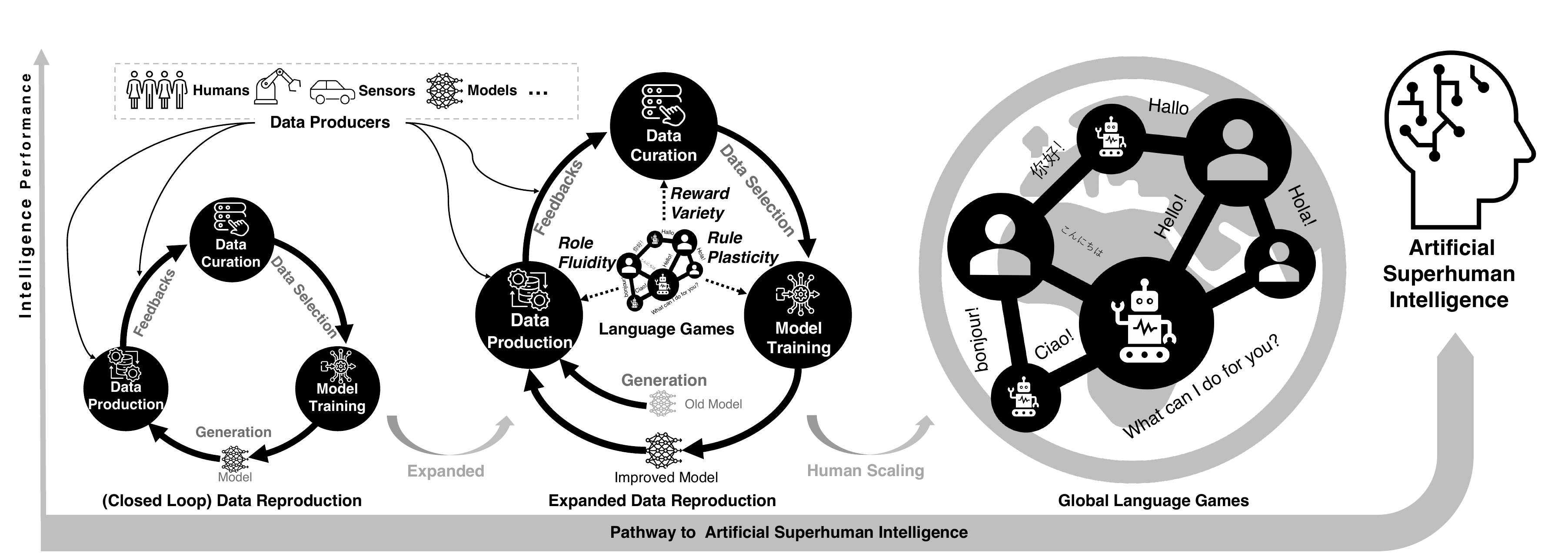}
\vspace{-15pt}
    \caption{An iterative data reproduction framework powered by language games progressively refines model capabilities toward artificial superhuman intelligence. From left to right: 
(1) \textbf{Data reproduction} via closed‐loop optimization, 
(2) \textbf{Expanded data reproduction via language games} with evolving roles, rules and rich rewards, 
(3) \textbf{Global language games} driving continual adaptation and surpassing human‐level capabilities.
}
    \label{fig:framework}
\end{figure*}

Escaping the data reproduction trap requires the transition to \textit{expanded data reproduction}, where models systematically transcend knowledge boundaries through environmental pressures. 
In this paper, we propose \textbf{language games}—dynamic linguistic interaction frameworks inspired by Wittgenstein’s view of meaning as use~\citep{wittgenstein1953philosophical}—as a means to operationalize expanded data reproduction.
\textbf{We argue that \emph{language games} are the pivotal mechanism for escaping the data reproduction trap in LLMs, thereby enabling open-ended conceptual exploration on the pathway to ASI.}
As stated in Fig.~\ref{fig:framework}, language games establish three mutually reinforcing mechanisms: \textit{role fluidity} enables models or agents to navigate diverse task spaces as both knowledge consumers and producers~\citep{subramaniam2025multiagent,shah2024agents,yang2021diverse}; \textit{reward variety} embodies the ``reward is enough'' hypothesis \citep{sutton1999reinforcement,silver2021reward,chambers2024learning} at metacognitive levels through pluralistic success criteria; \textit{Rule plasticity} sustains open-ended growth through linguistic environment remodeling—the capacity to reconfigure game constraints while maintaining learnable gradients~\citep{hughesposition,leibo2019autocurricula}. 
In addition, integrating reinforcement learning (RL) techniques into these language games can further enhance model proficiency, allowing them to internalize complex feedback signals and refine strategies in different domains~\citep{jaech2024openai,guo2025deepseek,wen2024reinforcing,su2025learn}. By constantly reshaping task roles, success metrics, and constraints, language games maintain an open-ended environment that fosters emergent behaviors critical to paving the way for ASI.

When language games are scaled up to the global human level, we establish a new sociotechnical ecosystem that acts as a data flywheel, enabling human-AI coevolution in both training and testing phases. 
With the rapid iteration of large models, costs are decreasing, and performance is steadily improving~\citep{brown2020language,achiam2023gpt,jaech2024openai,xiao2024densing}---especially given the explosive growth of open-source large models~\citep{guo2025deepseek,liu2024deepseek,touvron2023llama,dubey2024llama,yang2024qwen2}---thereby entering public consciousness, fostering technological equity and expanding societal penetration. 
As models engage in these language games, they internalize the deep structures of human reasoning while simultaneously exposing the boundaries of symbolic cognition~\citep{mumuni2025large}. 
At the same time, human participants discover the limits of their own conceptual frameworks as they grapple with AI-generated hypotheses~\citep{si2024can,hemmer2024complementarity,doshi2024generative,yan2024promises}.
This \emph{human scaling} integration not only amplifies the breadth and depth of data reproduction, but also accelerates the path to ASI by continuously refining and testing emerging model capabilities at a worldwide scale.

In summary, previous methods are constrained by static data distributions that focus on closed loop optimization. This shortsightedness restricts the emergence of deeply novel solutions and fails to support the open-ended exploration essential for transcending human-level intelligence. By contrast, language games introduce adaptive role assignments, diversified reward structures, and malleable game rules, thereby fostering a setting where emergent insights can thrive. Through these principles, we chart a course toward artificial superhuman intelligence built on ever-expanding data reproduction. In the following sections, we formalize data reproduction and language games (Sections~\ref{sec:data_reproduction}--\ref{sec:language_games}) and analyze the potential to scale them up to global human-level interactions in Section~\ref{sec:human_scaling}. We conclude in Section~\ref{sec:limitations} by discussing limitations and broader societal implications, then offer an alternative perspective on ASI in Section~\ref{sec:alternative_views}.

\section{Model Evolvement from Data Reproduction Perspective}
\label{sec:data_reproduction}

In this section, we examine how LLMs evolve via cycles of \emph{data reproduction}, and why certain trajectories risk stagnation (\emph{data reproduction trap}). We introduce \emph{expanded data reproduction} as a systematic strategy for ensuring ongoing capability growth.

\subsection{Data Reproduction}
The evolution of data-centric AI has transitioned through distinct eras, each shaped by how data are produced, selected, and used for model training, ultimately determining whether models remain confined to existing distributions or transcend them to acquire novel capabilities.

\textbf{Small-Model Era (Pre-2017).} Data production was typically narrow and task-specific, similar to ``small workshops'' creating specialized labeled sets (e.g. MNIST~\citep{lecun1999mnist}, ImageNet~\citep{deng2009imagenet}). Each reproduction cycle was short, limited by the availability of a new annotation for a fixed task.

\textbf{Large-Model Era (Post-2017).} Massive, factory-scale data collection (e.g., internet-scale text) fueled broad coverage\citep{radford2019language, brown2020language,achiam2023gpt}. However, even such large static corpora remain finite. After initial pretraining, subsequent data reproduction tends to be incremental, often relying on limited user feedback (e.g., RLHF) or fine-tuning sets, which can saturate quickly~\citep{brown2020language,ouyang2022training}.

In both eras, the key question remains: \emph{What new data is produced and how is it selected for training?} The quality, diversity, and novelty of \(X'\) directly affect whether the model becomes trapped in familiar distributions or breaks through to new conceptual territory. Based on this key question, we define the reproduction of data.

\begin{definition}[Data Reproduction]
Consider a model \(M_{\theta}\), such as an LLM, trained on an initial dataset \(D_0\). After deployment, the model and its environment generate new data \(X\) (e.g., user queries and responses, simulated interactions, sensor inputs). Then a subset \(X' \subseteq X\) is curated and used to retrain or fine-tune the model, producing an updated \(M_{\theta'}\). This process forms a closed loop where \(M_{\theta'}\) subsequently contributes to the generation of new data. Symbolically:

\resizebox{\linewidth}{!}{$
   M_{\theta}
   \;\xrightarrow[\substack{\text{Data Production}}]{\text{Generate }X}\;
   X
   \;\xrightarrow[\substack{\text{Data Curation}}]{\text{Feedback on }X}\;
   X^{\prime}
   \;\xrightarrow[\substack{\text{Model Training}}]{\text{Selected  }X^{\prime}}\;
   M_{\theta'}
   \; \circlearrowleft X.
$}
\end{definition}

This cyclical process, where each new iteration generates and re-absorbs data, constitutes \emph{data reproduction}. It can be viewed as a ``data metabolism mechanism'' that parallels the feedback loops seen in reinforcement learning or evolutionary algorithms \citep{sutton1999reinforcement,qian2024evolution,lee2025evolving}, but with a focus on the data \emph{itself} rather than solely on a model or policy. Over time, the model’s outputs reshape its future training distribution, creating an ongoing interplay between data generation and selection.

\subsection{Data Reproduction Trap}

\begin{definition}[Data Reproduction Trap]
Let \(p_{t}\) represent the effective distribution of curated data \(X'\) at iteration \(t\). Suppose we have a measure of distributional shift or divergence, \(\mathcal{D}(p_{t}, p_{t-1})\) (e.g., KL divergence or Wasserstein distance). 
We say the model falls into a \emph{data reproduction trap} if 
$$
  \mathcal{D}(p_{t}, p_{t-1}) \;\le\; \delta
  \quad 
  \text{for all}\ t \ge 1,
$$
where \(\delta\) is a small threshold over multiple iterations. Intuitively, the system re-ingests data drawn almost entirely from the same distribution it already mastered, yielding little genuine novelty. 
\end{definition}

There are many examples of this trap. One instance is homogeneous feedback, where the model interacts with users who provide short preference signals (like/dislike) biased toward ``safe'' or ``expected'' outputs. In such cases, the system naturally avoids riskier or more creative territory, stabilizing around a comfort zone and reproducing the same patterns with minimal exploration. Another example is self-reinforcing collapse, which can occur in extreme forms of self-play or self-training without external signals. If the curation mechanism prioritizes consistency over diversity, any stray drift toward an ``easy'' or degenerate distribution becomes amplified, collapsing model quality~\citep{shumailov2024ai,zhang2023selfcollapse,wang2024language,cheng2024self,pan2024spontaneous}. Although self-consistency methods such as majority voting have been shown to be effective in improving test-time performance~\citep{wang2022self}, they do not necessarily prevent convergence to limited distributional modes if the overall data pipeline lacks external corrective signals. In all these scenarios, the model becomes \emph{trapped} in a space of limited novelty, never receiving the impetus needed for deeper, more flexible intelligence.

\subsection{Expanded Data Reproduction}
\begin{definition}
Using the same measure of distributional shift \(\mathcal{D}(p_{t}, p_{t-1})\), we say the model exhibits \emph{expanded data reproduction} if there exists a positive threshold \(\Delta\) such that:
\[
  \frac{1}{T}\sum_{t=1}^{T} \mathcal{D}\bigl(p_{t}, p_{t-1}\bigr) \;>\; \Delta, 
\]
for multiple iterations \(T\). In other words, each round of data generation and selection injects sufficiently novel or out-of-distribution samples, ensuring a \emph{cumulative} expansion of the model’s knowledge distribution.
\end{definition} 

Expanded data reproduction avoids the data reproduction trap by systematically introducing content that the model has not yet mastered. This ensures \emph{persistent novelty}, as recurrent contact with unexplored domains prevents stagnation. It also enables \emph{adaptive curation}, by which curators or automated mechanisms selectively retain atypical but instructive samples to fill conceptual blind spots. Finally, it maintains \emph{balanced exploration and stability}, since novelty is targeted rather than random, allowing the model to learn from challenging data without spiraling into confusion. Over repeated cycles, these properties collectively drive open-ended learning by continuously shifting the effective data distribution away from what the model already dominates.

\section{Data Reproduction via Language Games}
\label{sec:language_games}  
In this section, we illustrate how \emph{language games}, defined at the level of linguistic interactions, serve as a powerful mechanism to achieve expanded data reproduction. We conclude with a comparative table that highlights the distinctions between language games and other data production approaches.

\begin{definition}[Language Games]
A \emph{language game} \(\mathcal{G}\) is defined in a linguistic space and specified by a tuple:
\[
  \mathcal{G} \;=\; \Bigl(\,\mathcal{A},\,\mathcal{S},\,\{\mathcal{U}_a\}_{a\in\mathcal{A}},\,R,\,\mathcal{E}\Bigr),
\]
where \(\mathcal{A}\) is a set of agents (LLMs, humans, specialized bots), \(\mathcal{S}\) is the shared semantic context (conversation history, tasks, partial solutions), \(\mathcal{U}_a\) is the action space for each agent \(a\) (e.g., statements, questions, instructions, critiques, codes), \(R\) is a \emph{reward function} reflecting correctness, creativity, ethical norms, or other criteria, \(\mathcal{E}\) defines how states evolve and how new roles, rules, or information enter the game.
\end{definition}

Because language games operate on open-ended dialogues, evolving objectives, and multi-agent roles, they naturally produce diverse trajectories of linguistic data. By continuously reshaping the incentives and constraints of participants, these games avoid reusing the same distributions and thus create new learning opportunities. The next subsections detail how these properties align with the concept of expanded data reproduction, preventing stagnation, and driving models toward broader, deeper capabilities.

\subsection{Link to Expanded Data Reproduction}

Language games provide a systematic mechanism for \emph{expanded data reproduction} by incorporating three core design principles that promote ongoing shifts in the model’s data distribution. These principles ensure a steady influx of novelty, diversity, and adaptive feedback, thereby averting common pitfalls such as distributional collapse or overfitting to uniform user preferences.

\paragraph{Role Fluidity.}
Agents in language games can dynamically assume different roles, such as teacher, student, protagonist, or critic \citep{chen2024persona,subramaniam2025multiagent}. This flexible assignment of perspectives multiplies the range of interactions and discourses, thereby producing a richer set of samples for curation. By shifting from question-generation to answer-verification, or from proposing hypotheses to challenging them, each agent introduces distinct patterns of reasoning and linguistic expression into the data pool. Over iterative fine-tuning, these varied viewpoints expand the model’s knowledge distribution, alleviating the risk of homogeneity and opening routes to entirely new domains.

\paragraph{Reward Variety.}
Instead of optimizing a single objective such as predictive accuracy or likelihood, language games incorporate multiple reward signals that can include logical consistency, domain fidelity, ethical compliance, creativity, or cultural sensitivity. This diversity of feedback offers a wider exploration space \citep{sutton1999reinforcement,silver2021reward}. When models encounter contradictory reward requirements—such as balancing factual correctness against imaginative novelty—they must generate data that negotiates these trade-offs. The resulting samples tend to be more varied and challenging, prompting curation pipelines to select examples that more effectively probe the model’s blind spots. Consequently, the repeated assimilation of such multifaceted data drives the model beyond conventional patterns, fueling sustained distributional shift.

\paragraph{Rule Plasticity.}
The environment of a language game remains malleable, allowing the introduction of new constraints, tasks, or cultural references over time. These ``rule injections'' prevent any single strategy or distribution from dominating and force the model to adapt to new situations~\citep{bengio2009curriculum,portelas2020automatic,leibo2019autocurricula,hughesposition}. By continuously altering the procedural or semantic parameters of the game, the system perpetuates novelty in both context and content. This perpetual renewal of game rules feeds back into data production, as interactions under newly evolved conditions generate off-distribution samples. With each retraining cycle, the model internalizes these emerging linguistic constructs, reinforcing \emph{expanded data reproduction} through iterative engagement with ever-changing challenges.

Together, these three principles ensure that language games remain an open-ended source of data. Novel roles and rewards stimulate exploration, while fluid rules enlarge the conceptual search space in which agents operate. The resulting distribution of interactions continually evolves, supplying fresh training signals that prevent the data reproduction trap and sustain long-term model growth.

\begin{table*}[ht!]
\centering
\vspace{-5pt}
\caption{Comparative analysis of data production paradigms. Language games uniquely combine high ratings across \emph{Data}, \emph{Feedback}, \emph{Improvement}, and \emph{Scalability}, positioning them as a strong candidate for catalyzing superhuman intelligence. See Table~\ref{tab:metric_definitions} in Appendix~\ref{app:eval} for metric definitions.}
\vspace{5pt}
\label{tab:data_production_comparison}
\resizebox{\linewidth}{!}{
\begin{tabular}{@{}lccccccccc@{}}
\toprule
\multirow{2}{*}{\textbf{Paradigm}} & 
\multicolumn{2}{c}{\textbf{Data}} & 
\multicolumn{2}{c}{\textbf{Feedback}} & 
\multicolumn{2}{c}{\textbf{Improvement}} & 
\multicolumn{2}{c}{\textbf{Scalability}} & 
\textbf{ASI Pathway} \\
\cmidrule(lr){2-3} \cmidrule(lr){4-5} \cmidrule(lr){6-7} \cmidrule(lr){8-9} \cmidrule(l){10-10}
 & Diversity & Volume & Richness & Quantity & Openendness & Emergence & Human & Autonomous & Viability \\
\midrule
Small Model Era& $\bullet$ & $\bullet$ & $\bullet$ & $\bullet$ & N/A & N/A & $\bullet\bullet$ & $\bullet$ & $\bullet$ \\
Large Model Era & $\bullet\bullet$ & $\bullet\bullet$ & $\bullet\bullet$ & $\bullet\bullet$ & $\bullet\bullet$ & $\bullet\bullet$ & $\bullet\bullet$ & $\bullet\bullet$ & $\bullet\bullet$ \\
Self-Play & $\bullet$ & $\bullet\bullet\bullet$ & $\bullet$ & $\bullet\bullet$ & $\bullet\bullet\bullet$ & $\bullet\bullet$ & $\bullet$ & $\bullet\bullet\bullet$ & $\bullet$ \\
\rowcolor{blue!10}
Language Games & $\bullet\bullet\bullet$ & $\bullet\bullet\bullet$ & $\bullet\bullet\bullet$ & $\bullet\bullet\bullet$ & $\bullet\bullet\bullet$ & $\bullet\bullet\bullet$ & $\bullet\bullet\bullet$ & $\bullet\bullet\bullet$ & $\bullet\bullet\bullet$\\
\bottomrule
\end{tabular}
}
\vspace{-5pt}
\end{table*}

\subsection{Examples of Language Games}
\label{subsec:lg_examples}

The \emph{Socratic Game}demonstrates how agents (human or AI) iteratively question assumptions to refine arguments and discover deeper insights~\citep{wilson2016socratic,schaul2024boundless}. A shared topic shapes the context \(\mathcal{S}\), while roles alternate between questioner and respondent. This swapping of perspectives promotes more diverse data production, since every participant must handle both critical inquiry and constructive explanation~\citep{vlastos1991socratic}. Reward signals evaluate logical coherence, originality, and compliance with evolving ethical or cultural norms, thereby expanding the model’s exposure to a wide range of reasoning patterns. Repeated transcripts containing nuanced arguments, counter-arguments, and hypothetical extensions feed back into training, pushing the model to integrate multi-perspective reasoning.

Other language games follow similar principles. Debate and negotiation sessions~\citep{du2023improvingdebate, irving2018aidebate, khan2024debating}, for instance, demonstrate how multiple agents (human-LLM teams or specialized bots) engage in policy-making or resource allocation, while an environment can inject new ethical frameworks or revised legal contexts. Collaborative role-playing may involve an AI assistant and a user co-authoring stories that undergo unforeseen ``plot twists,'' forcing the model to integrate novel stylistic and cultural elements. In simulated science laboratories, AI agents are prompted to propose and refine hypotheses, with occasional contradictory or surprising empirical data requiring inventive explanations. 

Each of these scenarios shares the open-ended, adaptive structure of language games, generating fresh, context-rich data that propels the model beyond a fixed knowledge distribution. By harnessing role fluidity, reward variety, and rule plasticity, language games build a robust scaffold for ongoing conceptual growth and critical reasoning—key ingredients for progressing toward superhuman intelligence.

\subsection{Comparisons With Other Data Production Modes}

Table~\ref{tab:data_production_comparison} contrasts language games with other paradigms by scoring each approach along dimensions crucial to \emph{expanded data reproduction}. In the Small Model Era (pre-2017), data production relied heavily on manual labeling and narrow tasks, limiting both diversity and volume of data. In the Large Model Era (2017--Now), large-scale web-scraped corpora and user-generated content greatly increased data coverage but remained mostly finite and static, often demanding fine-tuning strategies that risked reinforcing known patterns. Although self-play in a fixed ruleset can yield emergent solutions but lacks the cross-domain adaptability and open-ended scope required for genuine breakthroughs beyond a single problem space. By contrast, language games embed data production in a linguistically mediated, ever-evolving social context. Agents can draw on heterogeneous information streams, update reward criteria, and continuously modify role assignments or rule definitions. These features jointly foster a robust form of expanded data reproduction, as new tasks and contextual elements keep arising, thereby preventing the stagnation or overfitting seen in more static paradigms.

In summary, data reproduction is a unifying mechanism that explains how models improve by cycling through newly generated data. This mechanism appears in small-scale, large-scale, supervised, or self-play-based approaches, but in many cases, the data distribution saturates or collapses. Language games provide a more powerful framework for systematically overcoming such limitations. By orchestrating role fluidity, multiple reward signals, and plastic rules in an open-ended linguistic environment, language games offer a continuous supply of novel, complex data samples to refine the model. Through these repeated interactions, the model is driven beyond its initial distribution toward genuinely emergent capabilities that can ultimately approach or surpass human-level intelligence in multiple domains.

\section{Human Scaling: Global Language Games}
\label{sec:human_scaling}

When language games extend to a global scale, they evolve from localized experiments into self-reinforcing sociotechnical ecosystems. 
Driven by the rapid iteration of large models~\citep{brown2020language,achiam2023gpt,jaech2024openai,xiao2024densing}, decreasing costs and the explosive growth of open source communities~\citep{guo2025deepseek,liu2024deepseek,touvron2023llama,dubey2024llama,yang2024qwen2}, these ecosystems broaden the reach of AI beyond niche academic or corporate laboratories, fostering a participatory environment around the world.
This transformation creates a bidirectional evolutionary cycle between human communities and AI systems, driven by large-scale participation, cultural diversity, and dynamic feedback loops. 
In particular, the transition to a global scope amplifies the \emph{data reproduction} process: As more people engage with AI systems, both human and AI components continually refine each other’s cognitive boundaries.
The following discussion outlines how this ecosystem operates and provides empirical evidence of its potential to drive progress toward ASI.

\subsection{Global Language Games}

Global language games are realized through the interplay of billions of individual users, heterogeneous cultural settings, and continuously adapting LLMs. At the level of daily activity, people worldwide contribute open-ended dialogues, collaborative editing sessions, and strategic debates, creating large repositories of diverse cognitive patterns. Wikipedia exemplifies this dynamic: editors collectively refine articles, forming a long-term ``community proofreading mechanism''~\citep{halfaker2013rise}. As participation increases, cross-lingual and cross-regional exchanges give models exposure to varied cultural paradigms. Empirical work shows that multilingual models, when presented with tasks combining dialectical reasoning (often linked to Eastern traditions) and deductive inference (often tied to Western traditions), exhibit strengthened conceptual blending~\citep{norenzayan2006cultural}. By ingesting and synthesizing these diverse perspectives in real time, models update their parameters based on geographically dispersed user feedback. This continuous data reproduction loop enables them to internalize subtle cultural contexts and evolving semantic structures with unprecedented depth and scale.

\subsection{Dual-Phase Co-Evolution}

Within these global ecosystems, AI development proceeds through two intertwined phases of co-evolution. In the training phase, capability growth can accelerate nonlinearly as broader user participation surpasses critical thresholds of diversity. This phenomenon aligns with the \emph{cognitive diversity multiplier effect}, whereby cross-cultural and interdisciplinary disagreements prompt models to develop more general and robust representations. For instance, experimental results indicate that multilingual language games enhance meta-reasoning by forcing reconciliation of contradictory epistemic traditions~\citep{norenzayan2006cultural}. In the testing phase, large-scale deployment grounds model outputs in collectively validated truths. By exposing models to real-world scrutiny and diverse normative standards, distributional shift and value misalignment risks are mitigated. Findings from large-scale deliberative experiments suggest that consensus outputs generated through iterative debate can exhibit significantly higher robustness compared to decisions made within small groups~\citep{dafoe2020open}. Through these two phases, AI systems and human users incrementally refine each other’s cognitive processes, illustrating the dual-phase synergy of training acceleration and testing-grounded calibration.

\subsection{Pathways to Superhuman Intelligence}

The global scope of language games opens three promising pathways toward ASI. First, cross-cultural concept fusion allows models to merge distinct cognitive frameworks, such as integrating probabilistic reasoning with dialectical logic, leading to meta-cognitive abilities that surpass those found in any single tradition. Second, distributed proof markets employ game-theoretic incentives to crowdsource the verification of complex conjectures, significantly speeding up domains like mathematical theorem proving and scientific hypothesis testing~\citep{bergemann2019markets}. Third, consensus reality engineering leverages recursive debate to iteratively calibrate truth criteria, enabling collective oversight of knowledge claims in contentious areas such as climate science and ethical philosophy. Critically, each pathway is powered by large-scale human-model co-creation. Continuous feedback loops and data reproduction cycles allow models to outgrow the boundaries of human-expert performance, shifting AI research from a series of technical milestones to a sociotechnical process deeply embedded in communal intelligence.

\subsection{Adaptive Governance Frameworks}
The sheer scale and complexity of these global ecosystems call for innovative governance. Adaptive frameworks must balance incentives for rapid innovation against the need for robust ethical constraints, typically achieved through multiple layers of safeguards. At the individual level, differential privacy and federated learning protocols minimize personal data exposure while preserving collective benefits. System-level monitors continually check for deviations from evolving human values, triggering corrective actions whenever harmful content crosses predefined thresholds. Meanwhile, ecosystem-level mechanisms regulate the evolution of community rules, favoring decentralized negotiation over top-down directives. This vision follows recent work on pluralistic value alignment, where governance mechanisms evolve in tandem with the systems they guide~\citep{helbing2015sociotechnical}.

The shift to global language game ecosystems thus represents a move from designing AI tools to fostering a form of cognitive symbiosis. Models interwoven with global linguistic processes undergo constant scrutiny by a “Societal Turing Test,” facing continuous evaluation from diverse human viewpoints. At the same time, human participants expand their own conceptual reach when confronted with increasingly capable AI counterparts, mirroring the dialectical progression observed in localized language game experiments. This reciprocal evolution avoids the stagnation typical of static training pipelines and supports an open-ended developmental arc. The analyses of extensive collaborative platforms reinforce the view that such global co-evolution may represent the clearest route to a safe, scalable, and socially grounded ASI~\citep{dafoe2020open}.

\section{Limitations and Societal Challenges}
\label{sec:limitations}

Although language games offer a promising route toward more adaptive AI agents, realizing their full potential involves more than technical advances. Representational gaps, sociocultural inequities, ethical dilemmas, and epistemological questions complicate any large-scale deployment of AI in diverse human contexts. The following subsections highlight major challenges that may arise as LLMs move from localized tasks to proactive, interactive roles across global communities. We also offer concrete solutions such as decentralized identity protocols for mitigating oligopoly risks, localized review mechanisms for preventing reward hacking, and open research alliances for ensuring robust oversight.

\subsection{Expressive Limits and Information Loss}

Language provides a powerful medium for conceptual analysis and hypothesis testing, yet it remains an incomplete channel. Despite advances in multimodal techniques, textual and symbolic representations cannot capture every dimension of human sensation, such as tactile or emotional subtleties. Translating lived experiences into tokens often amplifies abstractions and introduces distortions. Although stronger prompt engineering and multimodal enhancements can narrow this gap, textual interfaces still risk omitting intangible aspects of reality. Existing research demonstrates that understanding and internalizing experiences differs significantly from simply reading about them \citep{bender2020climbing}. These limitations could restrict the forms of intelligence that evolve through language games, as certain categories of experiential knowledge remain beyond the scope of direct linguistic encoding. Continued work on integrated sensor modalities and embodied simulation may help close these expressive gaps.

\subsection{Knowledge Creation and Diffusion Risks}

Language games may transform the way knowledge is created, transmitted, and validated, because they foster continuous dialogues among users, AI agents, and changing contexts~\citep{robertson2024game}. This capacity for open interaction could spark innovative research trajectories or novel conceptual frameworks. However, questions of rigor arise when AI-generated ideas appear original but are not substantiated by external verification. Such illusions of discovery underscore the importance of epistemic checks and peer review. Although adaptive AI tutors hold the promise of individualized learning, hidden biases in underlying models may perpetuate existing educational disparities and reinforce social stratifications \citep{bowles2002schooling}. Deeper semantic modeling benefits information retrieval and content curation, yet there is a corresponding risk that privileged perspectives within the data overshadow minority viewpoints. Fostering robust epistemic diversity, along with reliable methods for confirming insights, requires deliberate governance and oversight. International collaborations and open research alliances can offer transparent platforms to evaluate AI-driven discoveries and maintain fair access to new knowledge~\citep{jumper2021highly}.

\subsection{Power Dynamics and Oligopoly Concerns}

Ideally, language games invite human participants to guide and enrich model evolution through multifaceted input. In practice, inequities and power imbalances can arise. Highly capable models, particularly those integrated into large-scale platforms, may gain oligopolistic advantages and shape collective discourse to discourage new entrants \citep{widder2024open}. Under such conditions, users risk being reduced to providers of raw data, with limited agency or mutual benefit \citep{greenbaum2025hidden}. The commercial exploitation of user creativity and attention can exacerbate disparities, effectively turning subjective experiences and personal engagement into monetizable assets \citep{zuboff2019age}. To mitigate such risks, decentralized identity protocols can anonymize participants in global language games enabling open collaboration beyond corporate boundaries. Regulatory bodies or alliances could enforce transparent disclosures of AI data usage, encourage multi-stakeholder participation, and establish fair access to computational resources. Without these safeguards, the concentration of AI resources and influence could restrict broader innovation, leaving marginalized communities with minimal impact on future AI directions.

\subsection{Overreliance and Manipulation Risks}

When language games scale up to personalized coaching, large-group ideation, or even collective problem-solving, the possibility of overreliance on AI systems becomes increasingly significant \citep{glickman2024human}. Users seeking practical answers or emotional support may compromise their own critical thinking, thereby transferring autonomy to opaque algorithmic processes. If the guiding reward systems prioritize retention or profit, malicious actors or misaligned agents might propagate sensational content or manipulate perceptions \citep{shneiderman2020human, xie2025discussions}. Individuals could be subtly directed or influenced by AI-driven narratives, raising concerns about informed consent and the gradual erosion of independent judgment. Cultural values, decision-making norms, and self-conceptions could shift in ways that are neither transparent nor easy to reverse, especially as users acclimate to AI outputs over long periods. Minimizing these vulnerabilities demands robust interpretability tools, continuous monitoring for manipulative behaviors, and legal frameworks that require accountability in algorithmic design.

\subsection{Cultural and Regulatory Complexity}

As language games expand across regions, institutions, and cultural settings, regulating them requires nuanced strategies. Attempts to define universal standards for what is ethically or socially acceptable may fail to account for local norms, while also risking forms of reward hacking in which models exploit misaligned objectives \citep{pan2024spontaneous} and reinforce certain viewpoints without broader contextual grounding. Such misalignments can overshadow minority perspectives and undermine balanced discourse. Geopolitical tensions also play a role: states and multinational corporations may vie for control over the policies guiding these games, reflecting broader patterns of global competition. Data sovereignty, privacy considerations, and the distribution of regulatory power remain contested. To address these concerns, multilingual and localized review mechanisms can detect and counter emergent reward hacking behaviors. 
These strategies call for ongoing audits, ethically aligned objective functions, and region-specific review panels that can adapt to different social contexts.

\subsection{Epistemological Realignment}

Dynamic and co-creative linguistic ecosystems challenge traditional notions of knowledge. Instead of simply reusing established data, language games allow AI to propose and reformulate concepts in ways that could blur the distinctions between legitimate discoveries and reworkings of existing information~\cite{schaul2024boundless}. Without appropriate validation or replication procedures, an echo chamber of superficial novelty may emerge. Ensuring that new insights are grounded in reliable sources demands iterated testing, structured peer review, and adherence to proven scientific principles. In addition, these ecosystems can shape how humans understand evidence and expertise. As users engage in immersive dialogues with AI, the boundaries of authoritative knowledge and the essence of meaning itself may be re-examined, for better or worse. Recognizing that language games act not only as neutral facilitators but also as active cultural drivers calls for multidisciplinary collaboration among ethicists, sociologists, and technologists.

\section{Alternative Views}
\label{sec:alternative_views}

A prominent alternative proposes that superhuman intelligence is based on \emph{embodied} interaction with physical environments, aligning with the perspectives of grounded world models \citep{lecun2022autointelligence,yang2024thinking}. Proponents might argue that language-centric systems lack the sensorimotor grounding for genuine understanding. Recent advances in robotics, such as self-improving manipulators \citep{bousmalis2023rt}, demonstrate how physical embodiment catalyzes causal reasoning through real-world feedback, echoing biological evolution \citep{pfeifer2007understanding}. Critics of purely linguistic methods emphasize three core divergences: (1) Physical survival objectives yield intrinsic rewards free from human bias; (2) Persisting spatio-temporal constraints mitigate hallucination; (3) Multimodal sensory integration provides cross-modal grounding missing in text-only systems. Although our language games excel at conceptual manipulation, advocates of embodied intelligence maintain that superhuman cognition requires a physics-anchored reality similar to that that shaped organic minds \citep{chiel2009brain}.

Yet this viewpoint need not compete directly with the language game paradigm. Rather, these two approaches can be complementary, as \emph{language games} excel at open-ended conceptual exploration and cultural fluency, while embodied systems ground those abstractions in sensorimotor experience. An agent can, for instance, describe its physical actions during real or simulated tasks as part of a language game, generating data that interlace verbal reasoning with environmental feedback. In turn, the model’s high-level symbolic inferences can guide more nuanced bodily interactions. This synergy fuses abstract and embodied learning, preventing hallucinations through real-world constraints while maintaining the creative potential of linguistic exploration. Although purely symbolic discovery—exemplified by mathematics—demonstrates the feasibility of non-embodied reasoning \citep{lakoff2000where}, integrating physical grounding may nonetheless enrich and stabilize the data reproduction cycles that drive superhuman intelligence.


\section{Conclusion}
\label{sec:conclusion}
This work positions language games as a transformative paradigm for achieving ASI through expanded data reproduction. By replacing static training loops with dynamic linguistic ecosystems, models escape the data reproduction trap imposed by human-centric optimization. The triad of role fluidity, reward variety, and rule plasticity ensures continuous distributional shift while maintaining learnable gradients—a critical balance absent in self-play or RLHF approaches. Global scaling amplifies these effects, transforming language games into a planetary-scale conceptual accelerator where cultural diversity and recursive self-improvement mutually reinforce capability growth.

Key challenges remain: mitigating representational losses in linguistic abstraction, preventing oligopolistic control, and preserving epistemic rigor amid AI-generated novelty. However, by framing ASI development as a coevolutionary process rather than a technical benchmark, language games align model advancement with collective human inquiry. This positions them not merely as training tools, but as infrastructure for symbiotic intelligence—a necessary step toward superhuman capabilities grounded in societal participation. Future work must address adaptive governance frameworks to ensure equitable growth as these ecosystems approach ASI thresholds.

\bibliography{references}
\bibliographystyle{icml2025}

\newpage
\appendix
\onecolumn
\section{Evaluation Metric Definitions}
\label{app:eval}

\begin{table*}[ht!]
\centering
\caption{Evaluation Metric Definitions.}
\label{tab:metric_definitions}
\resizebox{\linewidth}{!}{
\begin{tabular}{@{}lp{0.1\linewidth}p{0.75\linewidth}@{}}
\toprule
\textbf{Category} & \textbf{Metric} & \textbf{Definition} \\
\midrule
\rowcolor{gray!10}
Data & Diversity & Breadth of data coverage and novelty generation beyond training distribution boundaries. \\
\rowcolor{gray!10}
                & Volume & Rate of valid data generation per time unit, measuring system scalability. \\
\addlinespace
Feedback & Richness & Multi-source evaluation criteria including correctness, creativity etc. \\
                 & Quantity & Total amount of feedback provided to the system, impacting overall learning capacity. \\
\addlinespace
\rowcolor{gray!10}
Improvement & Openendness & Capacity for cross-domain knowledge exploration without predefined constraints. \\
\rowcolor{gray!10}
                  & Emergence & Frequency of spontaneous capability formation through self-organized learning. \\
\addlinespace
Scalability & Human & Efficiency gain in human-AI collaboration measured by knowledge transfer effectiveness. \\
            & Autonomous & Self-improvement capability without human intervention quantified by performance delta. \\
\addlinespace
\rowcolor{gray!10}
ASI Pathway & Viability & Probability assessment of achieving superhuman intelligence considering technical and theoretical factors. \\
\bottomrule
\end{tabular}
}
\end{table*}


\end{document}